\documentclass[runningheads]{llncs}
\usepackage[T1]{fontenc}

\usepackage[all, warning]{onlyamsmath}
\usepackage{amsmath,amssymb,balance,bm,color,mathtools,multirow, tabularx}
\usepackage{graphicx}
\usepackage{epstopdf}
\usepackage[table]{xcolor}
\usepackage[labelformat=simple]{subcaption}

\definecolor{Ired}{rgb}{0.8,0.0,0.0}
\definecolor{Iblue}{rgb}{0.0,0.0,0.7}
\definecolor{Igreen}{rgb}{0.0,0.6,0.0}
\definecolor{Gray}{gray}{0.9}

\newcommand{\Figref}[1]{{\bf Fig.\hspace{2pt}\ref{#1}}}

\newcommand{\Tabref}[1]{{\bf Table\hspace{2pt}\ref{#1}}}
\newcommand{\Eqref}[1]{Eq.\hspace{2pt}(\ref{#1})}
\newcommand{\Algorithmref}[1]{{\bf Alg.\hspace{2pt}\ref{#1}}}

\DeclareMathOperator*{\argmax}{arg\,max}

\usepackage[ruled,vlined,linesnumbered]{algorithm2e}

\makeatletter
{\list{}{\leftmargin=0pt 
		\itemsep=0pt
		\parskip=0pt
		\labelwidth\z@ \itemindent-\leftmargin
		}}%
{\endlist}
\makeatother
\begin{document}

\title{Multi-objectivization Relaxes Multi-funnel Structures in Single-objective NK-landscapes}
\titlerunning{Multi-objectivization Relaxes Single-objective NK-landscapes}
%

\author{Shoichiro Tanaka \and
Keiki Takadama\and
Hiroyuki Sato}
\authorrunning{S. Tanaka et al.}
%
\institute{Graduate School of Informatics and Engineering, \\The University of Electro-Communications, Tokyo, Japan \\
\email{stanaka@uec.ac.jp, keiki@inf.uec.ac.jp, h.sato@uec.ac.jp}}

\maketitle              
\begin{abstract}
This paper investigated the impacts of multi-objectivization on solving combinatorial single-objective $NK$-landscape problems with multiple funnel structures. Multi-objectivization re-formulates a single-objective target problem into a multi-objective problem with a helper problem to suppress the difficulty of the target problem. This paper analyzed the connectivity of two funnels involving global optima in the target and the helper $NK$-landscape problems via the Pareto local optimal solutions in the multi-objectivized problem. Experimental results showed that multi-objectivization connects the two funnels with global optima of the target and the helper problems as a single bridging domain consisting of the Pareto local optimal solutions. Also, this paper proposed an algorithm named the multi-objectivized local search (MOLS) that searched for the global optimum of the target problem from the global optimum of an artificially generated helper problem via the Pareto local optimal solutions. Experimental results showed that the proposed MOLS achieved a higher success rate of the target single-objective optimization than iterative local search algorithms on target $NK$-landscape problems with multiple funnels.
\end{abstract}

\section{Introduction}
The optimization performance of a search algorithm is determined by its behavior on the landscape of the target optimization problem. There are two main approaches to tackling optimization problems with rugged and complex landscapes. One is to develop search algorithms that work robustly on optimization problems with such landscapes. The other is to transform optimization problems with such landscapes into problems with simple landscapes. This paper focuses on the latter and multi-objectivization as one of the approaches \cite{knowles2001}. Multi-objectivization re-formulates an objective function ({\it target function}) of a single-objective target optimization problem into a multi-objective one. There are two types of multi-objectivization methods: those based on decomposition \cite{knowles2001} and those based on additional objectives \cite{jensen2004}. We limit the scope of this paper to the latter one, the multi-objectivization, by adding a new objective function ({\it helper function}) for solving the single-objective target problem. Previous works have reported that helper functions are effective for solving target problems with several difficult landscapes such as plateaus \cite{brockhoff2007} and multimodalities \cite{jensen2004,deb2010}. Multi-objectivization impacts the problem's difficulties since the fundamental change in the landscape \cite{garza2015}.

Among several landscape indicators, it has been known that local optima cause premature convergence of local search and harm global search methods such as evolutionary algorithms. On the other hand, recent landscape analysis research has shown that local optima distribution and connectivity affect the problem difficulties rather than the number of local optima \cite{thomson2017}. A cluster of local optima densely connected internally and sparsely connected externally is called \textit{funnel} and has recently attracted attention as a new indicator of the landscape. In an optimization problem with multiple funnels, a search algorithm may converge to a funnel including no global optimum. Even global search methods face difficulty in escaping from such a funnel. Indeed, in several optimization problem classes, funnels have been found to make the problem difficult to solve \cite{mcmenemy2018,ochoa2018,thomson2022}. To the best of our knowledge, the impact of multi-objectivization on the multi-funnel structure has yet to be studied.

In this paper, we investigate the impact of multi-objectivization on a single-objective target problem with a multi-funnel structure. We analyze the change in the landscape caused by the addition of helper problems. Each target and the helper problem is $NK$-landscape \cite{kauffman1987}, a combinatorial optimization problem with adjustable local optima and funnels. For combinations of the target and the helper $NK$-landscape problems with different problem parameters, we conduct a landscape analysis that observes the connectivity of their optima. In addition, we propose a search algorithm named the multi-objectivized local search (MOLS) that searches for the target problem from the global optimum of an artificially generated helper problem. We conduct an algorithm benchmark that compares the search performances of the proposed MOLS and the conventional iterated local search (ILS). Contributions of this paper are (1) to reveal how multi-objectivization relaxes multi-funnel structures in single-objective $NK$-landscape problems and (2) the proposal of MOLS that achieves better search performance on the target problem with multiple funnels than the conventional ILS.

\section{Single-objective Landscape}
\subsection{Single-objective Optimization}
Suppose we have a single-objective optimization problem with a pseudo-Boolean function $f:\{0, 1\}^N \rightarrow \bbbr$, which must be maximized. The search space is $X=\{0,1\}^N$, and the variable vector is $\bm x\in X$ which is a bit string of bit length $N$. The purpose of the single objective optimization is to search for the set of solutions $G:= \{\bm{x} \in X \mid \nexists \bm{x'} \in X; f(\bm{x})<f(\bm{x'})\}$ with the highest function value in the search space $X$. The solutions in $G$ are called global optima. 
The problems treated in this paper have only one global optimum $\bm{g}$.
In the search space $X$, local optima often interrupt convergence to a global optimum. Let $\mathcal{N}_{\epsilon}(\bm{x})= \{ \bm{x}' \in X \mid \bm{x}' \neq \bm{x}, d(\bm{x}, \bm{x}') \leq \epsilon \}$ be the $\epsilon$-neighborhood of solution $\bm{x}$, where $d(\bm{x}, \bm{x'})$ is the distance between the solutions $\bm{x}$ and $\bm{x'}$. The set of local optima $L$ of the single objective function $f$ can be defined as 
\begin{equation}
  L:=\{\bm{x} \in X \mid \nexists \bm{x}' \in \mathcal{N}_{\epsilon=1}(\bm{x});f(\bm{x}) < f(\bm{x}') \}.
\end{equation}
Note that $G \subseteq L$ from the above definitions.

\subsection{Basin of Attraction}
In this paper, the local search is the method of updating the search point to a solution with the highest function value among the neighborhoods of the search point. The pseudo-code of local search is shown in \Algorithmref{alg:ls}. The Basin of attraction can be defined as the set of solutions that converge to the same local optimum $\bm{l} \in L$ via local search. The convergence from a solution $\bm{x}$ to a local optimum $\bm{l} \in L$ by local search is denoted as ${\rm LS}(\bm{x})=\bm{l}$. The basin $B(\bm{l})$ of the local optimum $\bm{l}$ is defined as
\begin{equation}
  B(\bm{l}) := \{\bm{x}\in X \mid {\rm LS}(\bm{x})=\bm{l}\}.
  \label{eq:BoA}
\end{equation}
The search space $X$ can be divided into a finite number of basins $B(\bm{l})$ ($\bm l\in L$) when no plateau exists. For local search, the initial solution needs to exist in the optimal basin $B(\bm{g})$ to converge to a global optimum $\bm{g} \in G$. It is difficult to find global optimum $\bm{g} \in G$ by local search in an optimization problem with a tiny optimal basin $B(\bm{g})$.
\begin{algorithm}[t]
  \DontPrintSemicolon
  \LinesNotNumbered
  \setlength{\fboxsep}{1.5pt}
  \KwIn{initial solution $\bm{x}$, objective function $f$}
  \SetKwFunction{FunctionID}{LS}
  \SetKwProg{Fn}{Procedure }{:}{}
  \Fn{\FunctionID{$\bm{x}, f$}}{
    \While{$\bm x$ \rm{is updated}}
    {
        $\bm{x} \leftarrow \argmax_{\bm{x'}\in \mathcal{N}(\bm{x})}f(\bm{x})$
    }
    \KwRet{$\bm x$}
  }
  \caption{Local Search (best-improvement)}\label{alg:ls}
\end{algorithm}

\subsection{Local Optima Networks}
The funnel is defined using a local optima networks (LONs) model \cite{ochoa2008}, which represents the connectivity between local optima based on given search operators. LON is a weighted directed graph $G=(L, A)$, with the set of local optima $L$ as nodes and the set of transitions between local optima $A$ as arcs. This paper refers to directed edges as arcs. In this section, we describe a variant of the LON, the monotonic LON \cite{ochoa2017} and its arc, the perturbation arc \cite{verel2011}.
\begin{itemize}
\item \textbf{Monotonic LON: }
Monotonic LON (MLON) is a subgraph of LON. Nodes are local optima, and arcs are only ameliorate transitions between local optima by given search operators.
\vspace{0.3cm}
\item \textbf{Perturbation Arcs: }
The perturbation arc represents the transition based on the ILS operator. ILS is the trajectory-based algorithm and repeats local search and perturbation. The perturbation is a random bit-flip up to a fixed number of bits $\alpha$ for the solution $\bm x$. The pseudo-code for iterative local search is shown in \Algorithmref{alg:ils}. When a local optimum $\bm{l}\in L$ is updated to a local optimum $\bm{l'}\in L$, $\bm{l}$ and $\bm{l'}$ are connected by an arc $a(\bm{l}, \bm{l'})$. The arc weight $w(\bm{l}, \bm{l'})$ is the fraction of solutions in the $\alpha$-bit neighborhood $\mathcal{N}_{\alpha}(\bm{l})$ that converge to $\bm{l'}$ by the local search. When the perturbation arc $a(\bm{l}, \bm{l'})$ satisfies $f(\bm{l'})>f(\bm{l})$, it is called \textit{monotonic}.
\end{itemize}

\subsection{Funnel}
A monotonic LON with monotonic perturbation arcs has no cycles in the graph. Nodes in MLON involves a set of sink nodes $\hat{L}=\{\bm l\in L\mid \sum_{\bm{l'}\neq \bm{l}}w(\bm l, \bm{l'})=0\}$ that outgoing degree is zero. That is, each of sink nodes $\hat{L}$ has no transitions that improve the function value. When the local optimum $\bm{l}$ repeats the most probable transition, it converges to the sink point $\hat{\bm{l}}\in \hat{L}$.
This sequence of solutions is called a monotonic sequence\cite{ochoa2017} and is denoted ${\rm Path}(\bm{l})=\hat{\bm{l}}$. A set of local optima that converges to the same sink point $\hat{\bm l}$ via transitions is called funnel $F(\hat{\bm l})$ in this paper. The funnel $F(\hat{\bm l})$ is defined as
\begin{equation}
  F(\hat{\bm l}) := \{\bm{l}\in L \mid {\rm Path}(\bm{l})=\hat{\bm l}\}.
\end{equation}
Compared to \Eqref{eq:BoA}, a funnel is a basin of attraction at the level of local optima\cite{ochoa2017}. In single objective optimization, it has been reported that the number and depth of funnels affect the search \cite{mcmenemy2018,thomson2022}.

\begin{algorithm}[t]
  \DontPrintSemicolon
  \setlength{\fboxsep}{1.5pt}
  \KwIn{initial solution $\bm{x}$, objective function $f$}
  \SetKwFunction{FunctionID}{ILS}
  \SetKwFunction{Function}{LS}
  \SetKwFunction{ex}{Explored}
  \SetKwFunction{per}{Perturbation}
  \SetKwProg{Fn}{Procedure }{:}{}
  \Fn{\FunctionID{$\bm{x}, f$}}{
    $A \leftarrow \emptyset$\\
    $\bm l \leftarrow$ \Function ($\bm x$)\\
    \While{{\rm terminal criteria are not fulfilled}}
    {
      {
        $\bm x' \leftarrow $\per$(\bm l)$\\
      }
      \If{$\bm{x'}\notin A$}
      {
        $\bm l' \leftarrow $\Function$(\bm x')$\\
        $A \leftarrow A\cup\{\bm{x'}\}$\\
      }
      \If{$f(\bm l')>f(\bm l)$}
      {
        $\bm l \leftarrow \bm l'$\\
        $A \leftarrow \emptyset$\\
      }

    }
    \KwRet{$\bm x$}
  }
  \caption{Iterated Local Search}\label{alg:ils}
\end{algorithm}

\section{Multi-objective Landscape}
\subsection{Multi-objective Optimization}
Suppose that we have a multi-objective optimization problem with $M$ pseudo-Boolean functions $f_m:[0, 1]^N \rightarrow \bbbr$ ($m=1,\dotsc,M$), which must be maximized. The search space is $X=\{0,1\}^N$, and the variable vector is $\bm x\in X$ which is a bit string of bit length $N$. The purpose of multi-objective optimization is to search for a set of Pareto optimal solutions with an optimal trade-off relationship of multiple function values. For two solutions $\bm{x}, \bm{x'} \in X$, $\bm{x}$ dominates $\bm{x'}$ ($\bm{x} \succ \bm{x'}$) if
\begin{equation}
  \forall m \in \{1, \ldots, M\}; f_m(\bm{x}) \geq f_m(\bm{x'}) \ {\rm and} \ \exists m \in \{1, \ldots, M\}; f_m(\bm{x}) > f_m(\bm{x'}).
\end{equation}
A solution $\bm{x} \in X$ is said to be non-dominated in a solution set $S \subseteq X$ if no solution dominates $\bm{x}$ in $S$. The set of non-dominated solutions between neighborhoods is the Pareto local optimal solution set $PL:=\{\bm{x} \in \mathcal{X} \mid \nexists\bm{x'} \in \mathcal{N}_{\epsilon=1}(\bm{x}); \bm{x'} \succ \bm{x}\}$. The non-dominated solution set to the whole search space $X$ is the Pareto optimal solution set $P:=\{\bm{x} \in X \mid \nexists \bm{x'} \in X; \bm{x'} \succ \bm{x}\}$. Note that $P \subseteq PL$ from the above definitions.

\begin{figure}[t]
  \begin{center}
    \includegraphics[width=0.6\linewidth]{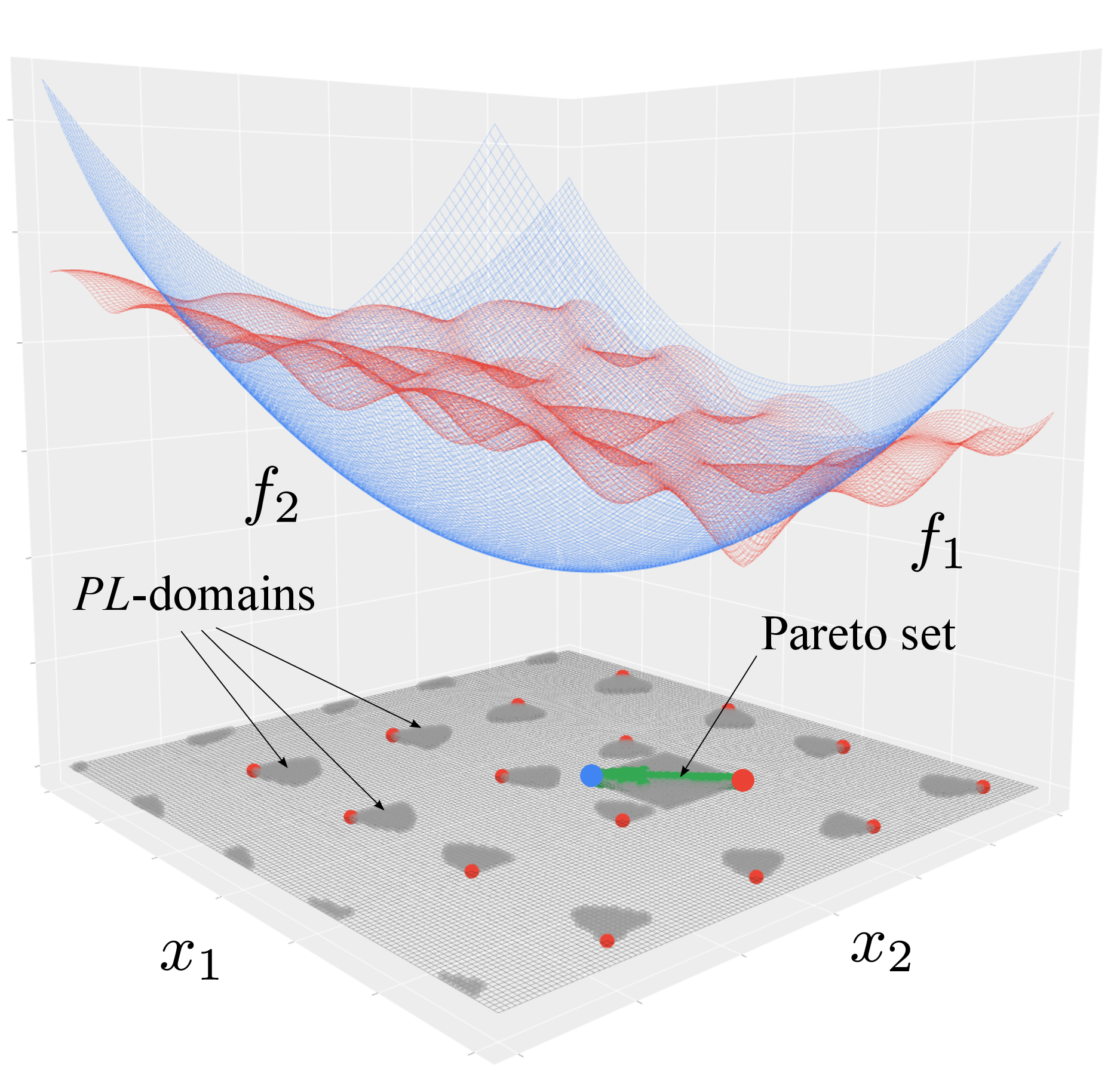}
    \caption{Visual example using a two-dimensional continuous search space for a two-objective minimization problem. The red circles are the local optima of $f_1$, the blue one is the global optimum of $f_2$, the green is the Pareto set of $\bm{f}=(f_1, f_2)$, and the gray are $PL$-domains, sets of the Pareto local optimal solutions.}
    \label{fig:2}
  \end{center}
\end{figure}

A local optimum $\bm{l}$ of the $m$-th objective function $f_m$ have higher $f_m(\bm{l})$ than its neighbors.
Thus, its neighbors do not dominate the $\bm{l}$.
Let $L_m$ be the local optima of the function $f_m$, the following relation holds: $\forall m \in \{1, \ldots, M\};L_m \subseteq PL$. This paper defines $PL$-\textit{domain} as a sub-domain consisting of a set of Pareto local optimal solutions in the search space. The search space $X$ can be divided into one or more $PL$-domains and non $PL$-domains. Since the Pareto optimal solutions and the local optimum of each objective function are Pareto local optimal solutions, each of them is involved in one of the $PL$-domains. 
For the sake of intuitive understanding, \Figref{fig:2} shows a two-dimensional continuous $x_1-x_2$ search space for a two-objective optimization problem $\bm{f}(x_1,x_2)=(f_1(x_1,x_2), f_2(x_1,x_2))$. Note that this paper deals with a discrete space, not the one shown in this figure. The red circles are the local optima for $f_1$, and the blue circle is the local optimum for $f_2$. The dark gray areas are $PL$-domains, and the green area is the Pareto set. We see that both local and Pareto optimal solutions exist in a $PL$-domain in this problem.

\subsection{Pareto Local Optimal Solutions Networks}
The Pareto local optimal solution network (PLOS-net) \cite{liefooghe2018} models the connectivity of Pareto local optimal solutions in combinatorial search space. PLOS-net is an unweighted undirected graph $G=(PL, E)$ with the Pareto local optimal solution set $PL$ as nodes and the transitions between the solutions $E$ as edges. When Pareto local optimal solutions $\bm{pl}, \bm{pl'} \in PL$ are adjacent to each other $\bm{pl} \in \mathcal{N}(\bm{pl'})$, $\bm{pl}$ and $\bm{pl'}$ are connected by an edge $E(\bm{pl}, \bm{pl'})$. PLOS-net has one or more components corresponding to $PL$-domains.
This paper focuses on the $NK$-landscape problem \cite{kauffman1987} and the $MNK$-landscape problem \cite{aguirre2004}.
\section{NK- and MNK-landscape problems}
\subsection{NK-landscape problem}
Solution is a bit string $\bm{x}=(x_1,x_2,\dotsc,x_{N})\in\{0,1\}^N$ of bit length $N$. The search space size $|X|$ is $2^{N}$. The objective function of the $NK$-landscape problem that must be maximized is a pseudo-Boolean function and is defined as
\begin{equation}
f^{NK}(\bm{x}) = \frac{1}{N}\sum^{N}_{j=1}g_j(x_{j}, x_{j,1}, x_{j,2}, \ldots, x_{j,K}),
\end{equation}
where $g_j$ ($j=1,\dotsc,N$) are the sub-functions, $x_{j, 1}, x_{j, 2}, \ldots, x_{j, K}$ are the randomly selected co-variables for the $j$-th sub-function $g_j$ and $K$ is the number of co-variables. In this paper, all instances of $NK$-landscape problems have only one global optimum $\bm{g}$ and no plateau. The $NK$-landscape problem framework is tunable from a smooth unimodal landscape to a rugged multimodal landscape as $K$ increases from $0$ to $N$. In general, as $K$ increases, the difficulty increases. The function value $f^{NK}(\bm{x})$ is calculated by variable relationships determined by co-variables and a random number table. By changing the variable relationships randomly determined by co-variables or the table, we can generate different instances with the same problem parameters $N$ and $K$.

\subsection{MNK-landscape problem}
The $MNK$-landscape problem is an extension of the $NK$-landscape one and involves $M$ objective functions with the same solution space $X$. The objective functions of the $MNK$-landscape problem are defined as
\begin{align}
  \bm{f}^{MNK}(\bm{x}) = (f_{1}^{NK}(\bm{x}), \ldots, f_M^{NK}(\bm{x})),
\end{align}
where
\begin{align}
  f_i^{NK}(\bm x) = \frac{1}{N}\sum^{N}_{j=1}g_{i,j}(x_j, x_{i,j,1}, \ldots, x_{i,j,K_i})~~(i=1,\dotsc,M). \label{f_sub}
\end{align}
Each objective function $f_i^{NK}$ is an $NK$-landscape problem and consists of $N$ sub-functions $g_{i,j}$ $(j=1,\ldots,N)$, and its each sub-function $g_{i,j}$ takes variable $x_j$ and $K_i$ co-variables. In this paper, $NK$-landscapes, which constitutes $MNK$-landscape, are uncorrelated.
\begin{figure*}[t!]
  \begin{center}
    \includegraphics[width=\linewidth]{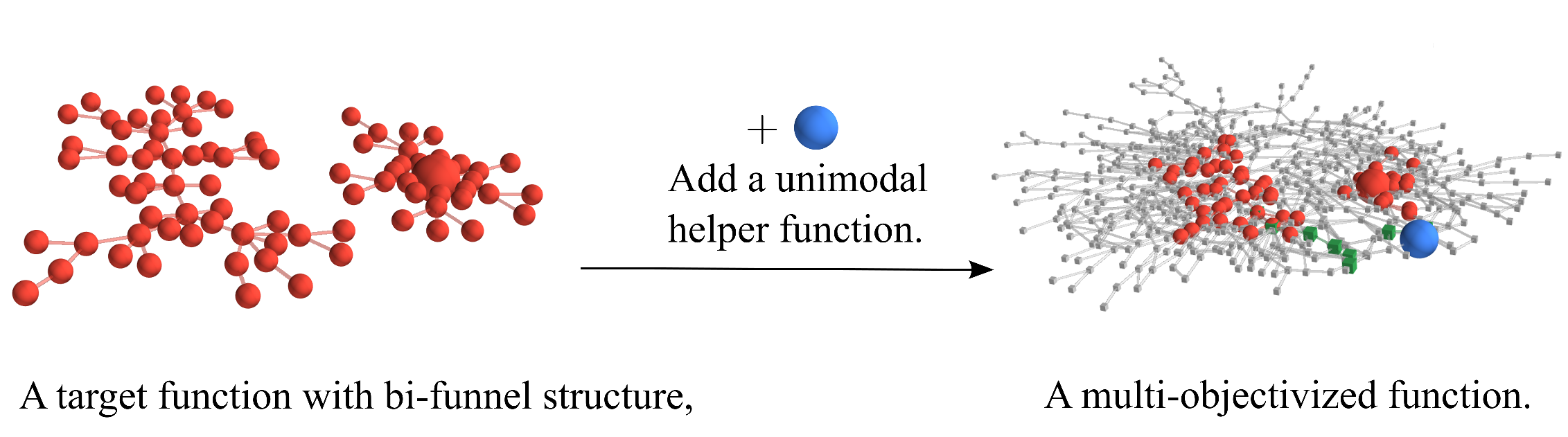}
    \caption{Change in the landscape with the addition of a unimodal helper objective. Red circles are the local optima of target objective $f_{\rm target}$, and the blue one is the global optimum of helper $f_{\rm helper}$, green ones are the Pareto set of $\bm{f}=(f_{\rm target}, f_{\rm helper})$, and gray ones are $PL$-domain, the set of Pareto local optimal solutions.}
    \label{fig:3}
  \end{center}
\end{figure*}
\section{Proposed Method: Multi-Objectivized Local Search}
\subsection{Motivation}
As the variable relationships $K$ increase, local optima generally increase. As a result, the size of the optimal basin $B(\bm g)$ and funnel $F(\bm g)$, including a global optimum $\bm g$, becomes small. This is one of the intuitive reasons why optimization problems become more difficult to solve. In contrast, a study \cite{liefooghe2018} using PLOS-net suggests that the $PL$-domain appears as one large domain independent of variable relationship strength. 

\Figref{fig:3} shows the change in the landscape when a multimodal instance of a single objective target $NK$-landscape $f_{\rm target}$ is multi-objectivized with a unimodal instance of an additional helper one $f_{\rm helper}$. The red, blue, and gray graphs represent the LON of the target $f_{\rm target}$, the helper $f_{\rm helper}$, and the PLOS-net of $\bm{f}=(f_{\rm target}, f_{\rm helper})$, respectively, while the green nodes represent the Pareto set. From the left side of \Figref{fig:3}, we see that the red LON of the target $f_{\rm target}$ has two red components, funnels. From the right side of \Figref{fig:3}, the target bi-funnel structure in red is transformed into a landscape with one huge $PL$-domain in gray by adding the helper function $f_{\rm helper}$. If the search space has a multi-funnel structure, search algorithms may converge to a funnel with no global optimum. On the other hand, if the $PL$-domain can connect the multiple funnels, the helper function enables transitions between the multiple funnels via Pareto local optimal solutions. Multi-objectivization aggregates the multiple funnels into a $PL$-domain and suppresses the search difficulty due to the multi-funnel structures.

\begin{algorithm}[t]
  \fontsize{8.7pt}{9pt}\selectfont
  \DontPrintSemicolon
  \setlength{\fboxsep}{1.5pt}
  \KwIn{objective function $f_{\rm target}$, helper function $f_{\rm helper}$}
  \SetKwFunction{ex}{Explored}
  \SetKwFunction{FunctionID}{MOLS}
  \SetKwProg{Fn}{Procedure }{:}{}
  \Fn{\FunctionID{$f_{\rm target}, f_{\rm helper}$}}{
    $\bm{x} \leftarrow$ a global optimum of $f_{\rm helper}$ \label{alg_ln: init}\\
    $\bm{f} \leftarrow (f_{\rm target}, f_{\rm helper})$\\
    $A \leftarrow \{\bm{x}\}$ \label{alg_ln: init archive}\\
    \While{{\rm terminal criteria are not fulfilled} \label{alg_ln: focus x}}
    {
      $\mathcal{N}^+ \leftarrow \emptyset$\label{alg_ln: make empty}\\
      \ForEach{$\bm{x'} \in \mathcal{N}_{\epsilon=1}(\bm{x})$ \label{alg_ln: try neighors}}
      {
        \If{$\bm x \nsucc \bm x' \land \bm{x'} \notin A$ \label{alg_ln: collect candidate neighors}}
        {
          $\mathcal{N}^+ \leftarrow \mathcal{N}^+\cup\{\bm x'\}$\\
        }
      }
      $\bm{x} \leftarrow \argmax_{\bm{x'}\in \mathcal{N}^+}f_{\rm target}(\bm{x})$\label{alg_ln: check candidate neighors}\\
      $A \leftarrow A \cup \{\bm{x}\}$\label{alg_ln: update}\\
    }
    $\bm{x} \leftarrow \argmax_{\bm{x}\in A}f_{\rm target}(\bm{x})$\label{alg_ln: output}\\
    \KwRet{$\bm{x}$}
    }
  \caption{Proposed Multi-Objectivized Local Search (MOLS)}\label{alg:mols}
\end{algorithm}  

\subsection{Algorithm}
In this paper, we propose an algorithm that searches for a global optimum of a target function $f_{\rm target}$ using a multi-objective problem consisting of the target function $f_{\rm target}$ and an additional helper function $f_{\rm helper}$. 

\Algorithmref{alg:mols} shows the pseudo-code of the proposed search algorithm named the multi-objectivized local search (MOLS). The input is the target function $f_{\rm target}$ with $N$ variables and a helper function $f_{\rm helper}$, which is a randomly generated unimodal instance of $NK$-landscape problem with $N$ variables and $K=0$ co-variables.
It is not always appropriate to choose such a unimodal function that is independent of the target function for the helper function.
However, there are two reasons for this design.
First, in a black-box scenario, providing helper functions that are not independent of the target function, i.e., correlated in some respect, is not easy.
Therefore, we decided to generate independent helper functions as a first step.
Second, the reason for using unimodal helper functions is that it is easy to find a global optimum.
This algorithm expects transitions between global optima and requires knowledge of the global optimum of the helper function.
At line \ref{alg_ln: init}, we set an initial solution $\bm x$.
The proposed MOLS does not start with a randomly generated initial solution such as conventional search algorithms. The proposed MOLS starts with the global optimum $\bm{g}_{\rm helper}$ of the helper function $f_{\rm helper}$. The computational complexity of searching for the global solution $\bm{g}_{\rm helper}$ of the helper function $f_{\rm helper}$ is $\mathcal{O}(N)$. Since the initial solution of MOLS is a global optimum $\bm{g}_{\rm helper}\in G_{\rm helper}$ of the helper function $f_{\rm helper}$ and a Pareto local optimal solution at the same time, MOLS searches only within the $PL$-domain where $\bm{g}_{\rm helper}$ exists. At line \ref{alg_ln: init archive}, the initial solution joins to the archive $A$, and the repetition started from line \ref{alg_ln: focus x} searches the neighborhoods of the current solution $\bm x$. At line \ref{alg_ln: make empty}, the set of candidate neighborhoods $\mathcal{N}^+$ is prepared as the empty set. In the repetition started from line \ref{alg_ln: try neighors}, we collect candidate neighborhoods $\mathcal{N}^+$ from the neighborhoods $\mathcal{N}_{\epsilon=1}(\bm x)$ of the current solution $\bm x$. The neighborhoods $\mathcal{N}_{\epsilon=1}(\bm x)$ not dominated by the current solution $\bm x$ and not in the archive $A$ joins to the candidate neighborhoods $\mathcal{N}^+$. In the repetition started from line \ref{alg_ln: check candidate neighors}, we take the best solution in the candidate neighborhoods $\mathcal{N}^+$ in the viewpoint of the target objective function value $f_{\rm target}$ as the next focal solution $\bm x$. At line \ref{alg_ln: update}, we put it to the archive $A$. We then repeat the above process.

In this way, the proposed MOLS searches the global optimum $\bm{g}_{\rm target}$ of the target objective function $f_{\rm target}$ from the global optimum $\bm{g}_{\rm helper}$ of the helper objective function $f_{\rm helper}$. If global optima of both target and helper functions exist in the same $PL$-domain, MOLS can reach the global optimum $\bm{g}_{\rm target}$ of the target function $f_{\rm target}$ in principle.

\section{Experimental Setup}
We conducted two experiments in this paper. One is (1) the landscape analysis, and the other is (2) the algorithm benchmarks.
\subsection{Landscape Analysis}
In the landscape analysis, we quantitatively investigated the landscape changes caused by multi-objectivization. 

We generated independent $10$ instances for each $NK$-landscape problem setting with $N=18, K\in\{0,2,4,6,8,10\}$, for a total of $60$ instances. The $60$ instances of $NK$-landscape problems were quantitatively evaluated using the metrics shown in \Tabref{tab:1}. We obtained the funnel using MLONs with $\alpha=2$.

As combinations in the $60$ instances of single-objective $NK$-landscape problems generated above, we generated $\binom{60}{M=2}=1,770$ instances of $MNK$-landscape problem setting with $M=2$, $N=18$, and $K_1, K_2\in\{0, 2, 4, 6, 8, 10\}$. The $1,770$ instances of $MNK$-landscape problems are quantitatively evaluated using the metrics shown in \Tabref{tab:2}. We obtained the $PL$-domain using the PLOS-net \cite{liefooghe2018}. Graph features are obtained for the only component corresponding to the optimal $PL$-domain.
\begin{table}[t!]
  \caption{Metrics for single-objective landscapes}
  \label{tab:1}
  \begin{center}
  {%
    \begin{tabular}{l|l}
    \hline
    Metrics                          & Description                                                                                     \\ \hline \hline
    $\mathcal{B}$                    & The number of the basin of attraction.                                                          \\ \hline
    $\mathcal{F}$                    & The number of the funnels.                                                                      \\ \hline
    $\mathcal{B}^*/\mathcal{X}$      & The relative size of the optimal basin to the search space $|B(\bm{g})|/|X|$                        \\ \hline
    $\mathcal{F}^*/\mathcal{L}$      & The relative size of the optimal funnel to the search space $|F(\bm{g})|/|L|$.                      \\ \hline
    \end{tabular}
  }
  \end{center}
  \vspace{-0.7cm}
\end{table}

\begin{table}[t]
  \caption{Metrics for multi-objective landscapes}
  \label{tab:2}
  \begin{center}
  {%
    \begin{tabular}{l|l}
    \hline
    Metrics                      & Description                                                                              \\ \hline \hline
    $\mathcal{PL}/\mathcal{X}$   & The ratio of the Pareto local optimal solutions $PL$ to the search space.       \\ \hline
    $\mathcal{D}$                & The number of $PL$-domains.                                                              \\ \hline
    $\%\mathcal{D}^*$            & The existence probability of optimal $PL$-domain $D^*$.                                     \\ \hline
    $\mathcal{D}^*/\mathcal{X}$  & The relative size of $D^*$ to the search space $X$.                                      \\ \hline
    $\mathcal{D}^*/\mathcal{PL}$ & The relative size of $D^*$ to $PL$.                                                      \\ \hline
    $path^*$                     & The length of shortest path $path^*$ between global optima in $D^*$.                     \\ \hline
    $\mathcal{P}/path^*$         & The fraction of the Pareto optimal solutions in $path^*$.                                    \\ \hline
    $deg\_\mathcal{D}^*$         & The average degree of $D^*$.                                                             \\ \hline
    $deg\_path^*$                & The average degree of $path^*$.                                                          \\ \hline
    \end{tabular}
  }
  \end{center}
  \vspace{-0.7cm}
\end{table}

\begin{table}[t]
  \caption{Metrics of search performance}
  \label{tab:3}
  \begin{center}
  {%
    \begin{tabular}{l|l}
    \hline
    Metrics             & Description                               \\ \hline \hline
    $\mathcal{SR}$      & The fraction of trials in which a global optimum was found in $1,000$ trials. \\ \hline
    $\mathcal{ST}$      & The number of function evaluations when the global optimum is found.          \\ \hline
    $\mathcal{CT}$      & The number of function evaluations at the end of the search.                  \\ \hline
    \end{tabular}
  }
  \end{center}
  \vspace{-0.7cm}
\end{table}

\subsection{Algorithm Benchmarks}
In the algorithm benchmark, we quantitatively assessed the optimization performance of the target function.

We compared three search algorithms: the proposed MOLS and the two conventional ILSs with different perturbation strengths $\alpha \in \{2, 3\}$, which gives perturbations of more than $2$-bits and less than $\alpha$-bits. We considered that ILS has converged if the focal solution has not been updated after performing a local search from all possible solutions obtained by perturbation. ILS terminated when it either converged or reached the maximum number of function evaluations $FE_{max} =N^2\times100$. The proposed MOLS terminated if there was no unexplored Pareto local optimal solution in the neighborhood of the focal solution or reached the maximum number of function evaluations $FE_{max} =N^2\times100$. 

We used independent $30$ instances for each $NK$-landscape problem setting with $N=18, K\in\{2,4,6,8,10\}$, for a total of $150$ instances. For each instance, each search algorithm was executed 1,000 times independently.
We used three metrics shown in \Tabref{tab:3}. Note that $\mathcal{ST}$, the number of function evaluations until the global solution was found, was the average only among the successful runs.

\section{Experimental Results and Discussions}
\subsection{Landscape Analysis}
\begin{figure*}[t]
  \begin{minipage}{0.325\textwidth}
    \begin{center}
      \includegraphics[width=1.0\textwidth,clip]{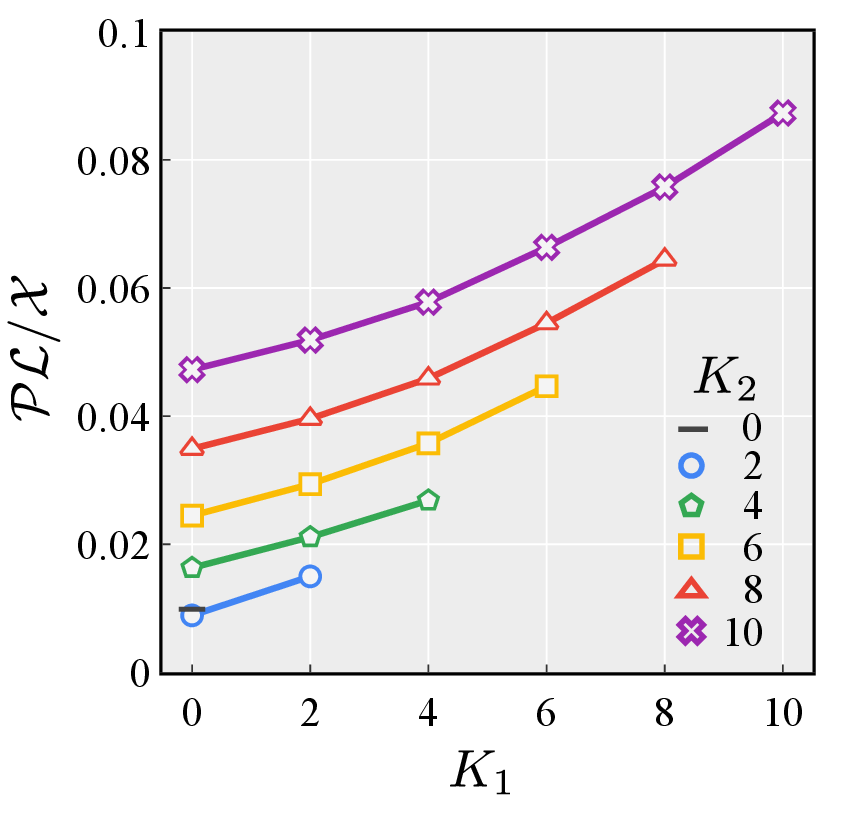}
      \\\vspace{-0.1cm}
      \subcaption{The ratio of $PL$.}
      \label{fig:4a}
    \end{center}
  \end{minipage}
  \begin{minipage}{0.325\textwidth}
    \begin{center}
      \includegraphics[width=1.0\textwidth,clip]{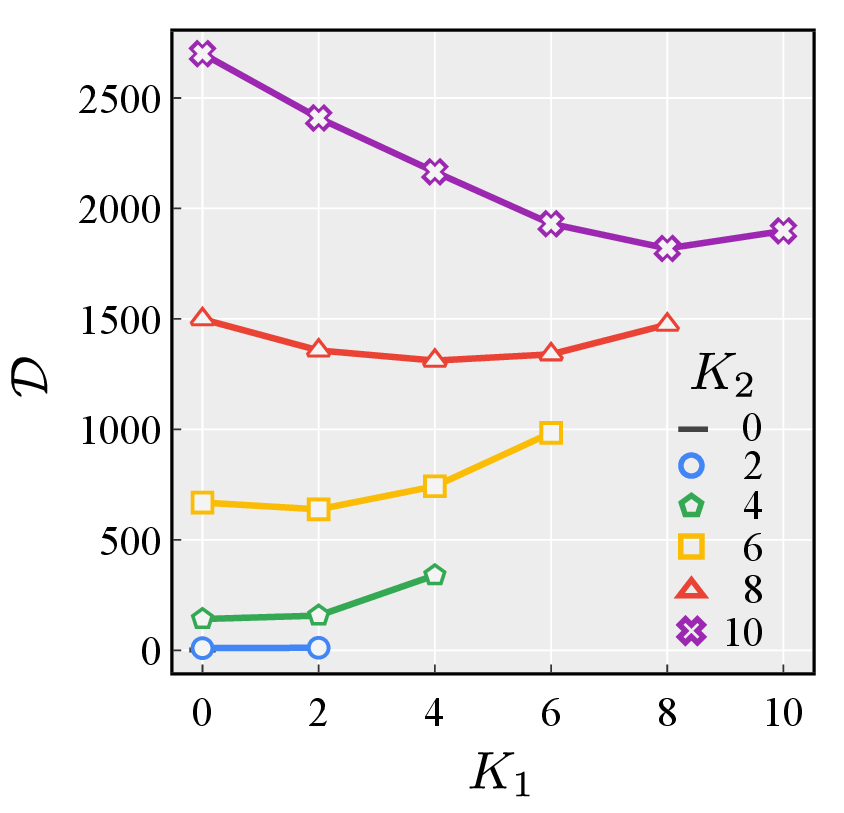}
      \\\vspace{-0.1cm}
      \subcaption{The number of $D$. }
      \label{fig:4b}
    \end{center}
  \end{minipage}
  \begin{minipage}{0.325\textwidth}
    \begin{center}
      \includegraphics[width=1.0\textwidth,clip]{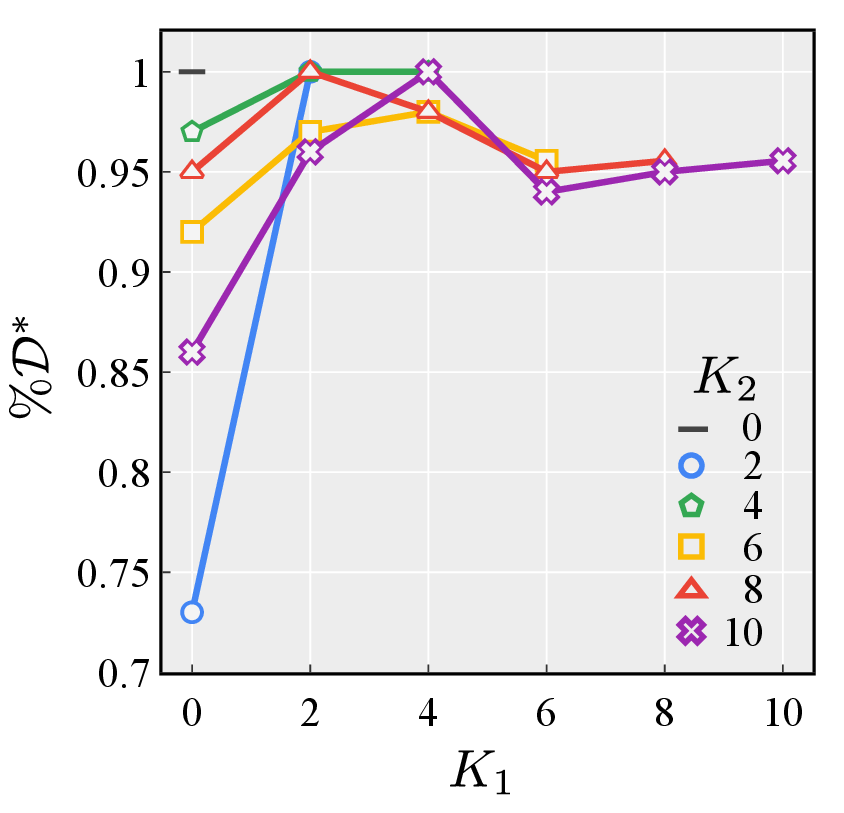}
      \\\vspace{-0.1cm}
      \subcaption{The probability of $D^*$.}
      \label{fig:4c}
    \end{center}
  \end{minipage}

  \begin{minipage}{0.325\textwidth}
    \begin{center}
      \includegraphics[width=1.0\textwidth,clip]{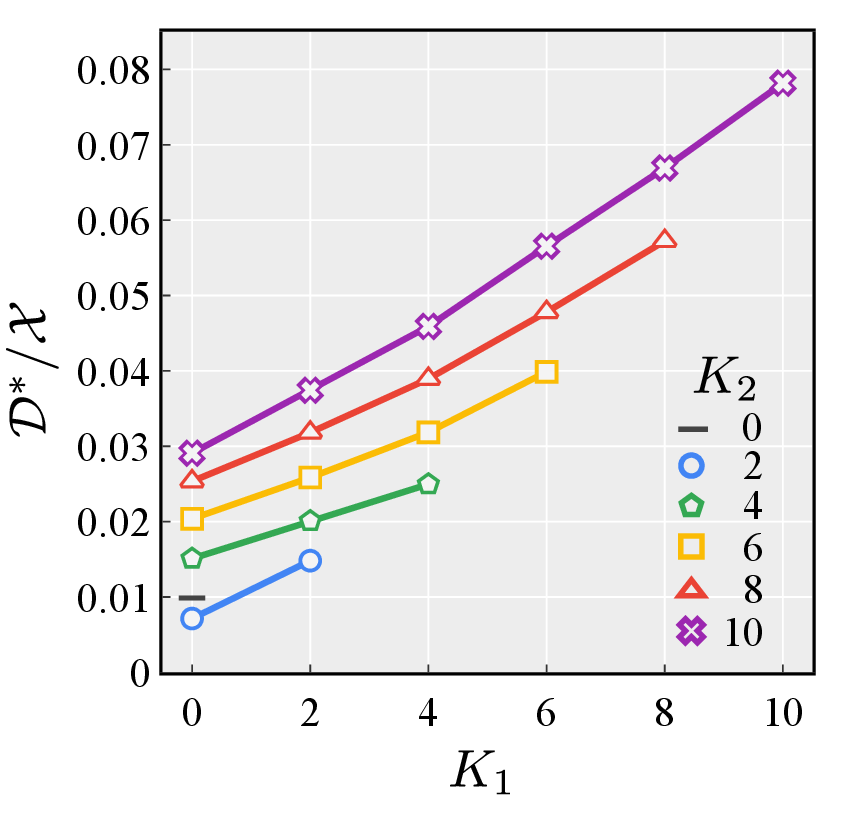}
      \\\vspace{-0.1cm}
      \subcaption{The size of $D^*$.}
      \label{fig:4d}
    \end{center}
  \end{minipage}
  \begin{minipage}{0.325\textwidth}
    \begin{center}
      \includegraphics[width=1.0\textwidth,clip]{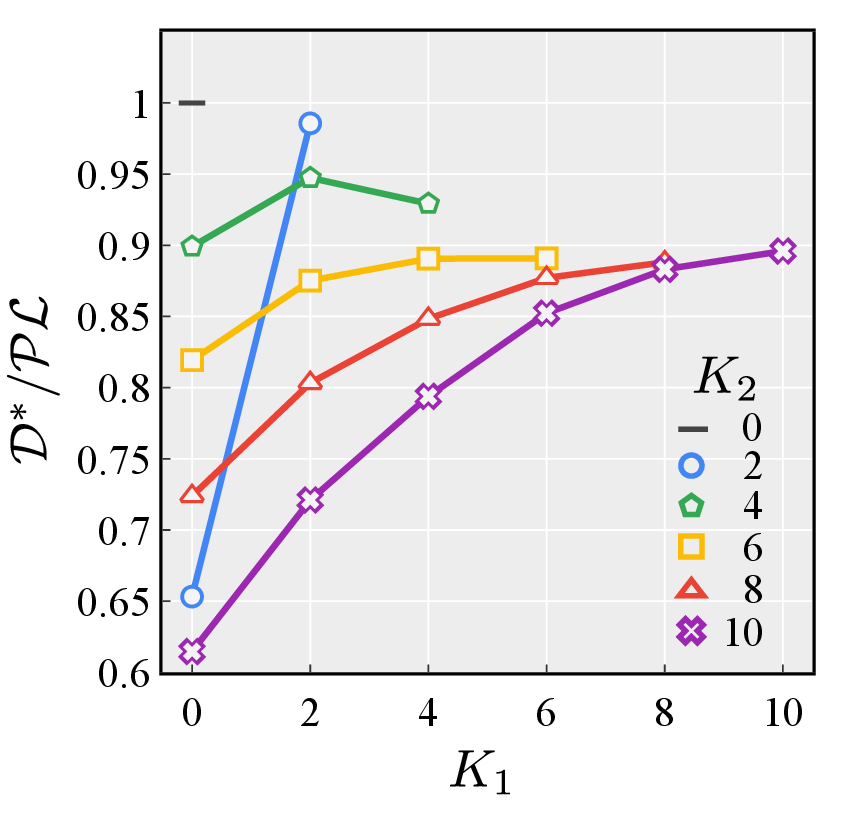}
      \\\vspace{-0.1cm}
      \subcaption{The ratio of $D^*$ to $PL$.}
      \label{fig:4e}
    \end{center}
  \end{minipage}
  \begin{minipage}{0.325\textwidth}
    \begin{center}
      \includegraphics[width=1.0\textwidth,clip]{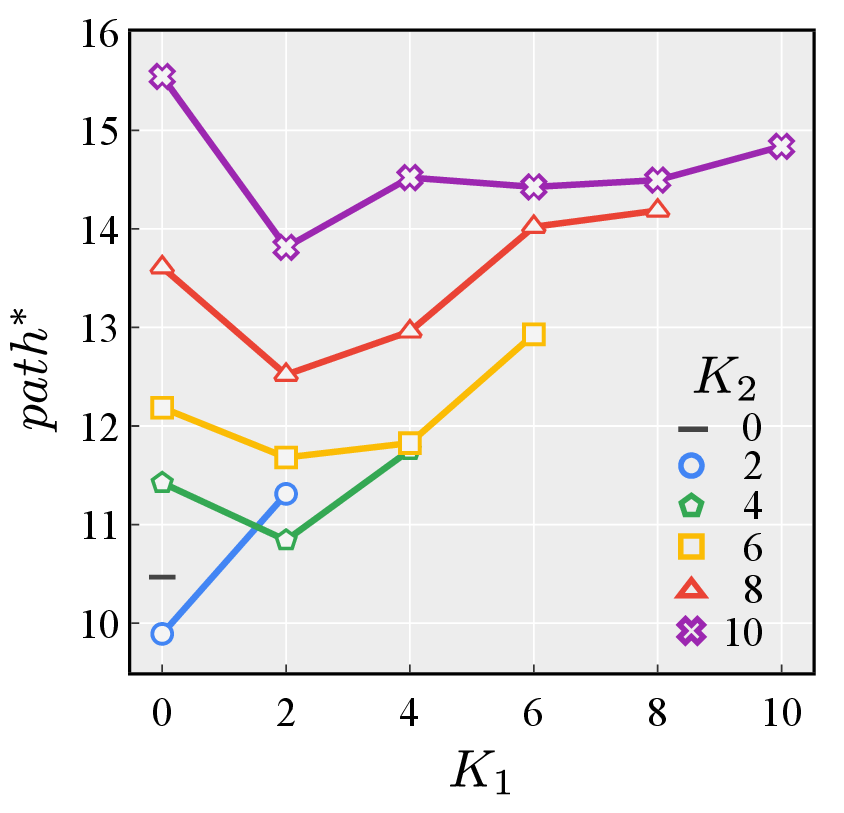}
      \\\vspace{-0.1cm}
      \subcaption{The length of $path^*$.}
      \label{fig:4f}
    \end{center}
  \end{minipage}

  \begin{minipage}{0.325\textwidth}
    \begin{center}
      \includegraphics[width=1.0\textwidth,clip]{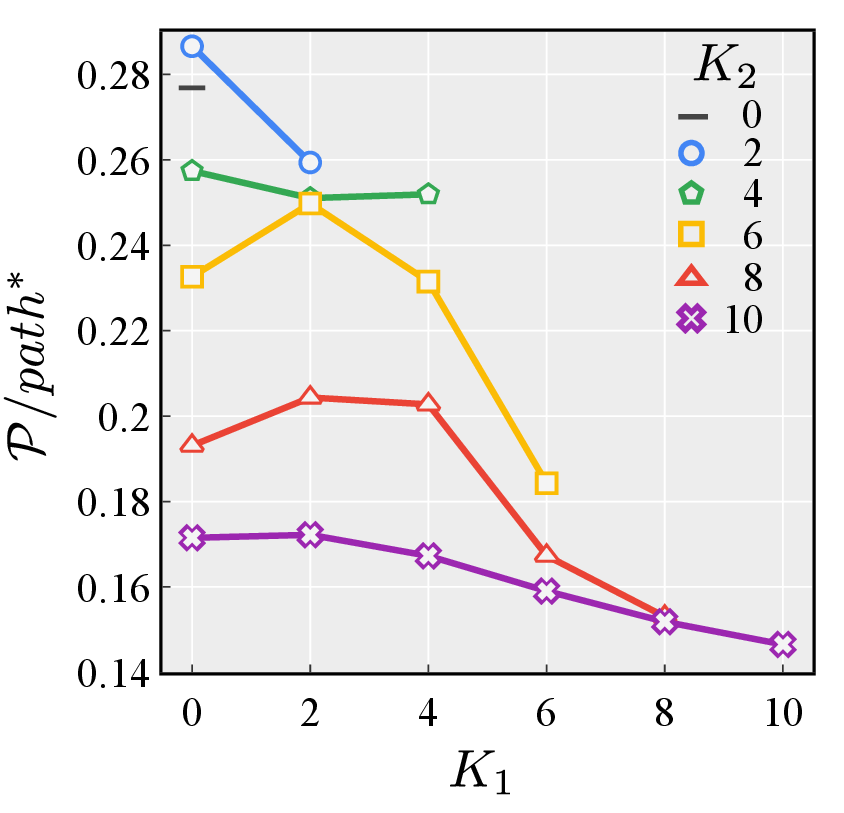}
      \\\vspace{-0.1cm}
      \subcaption{The fraction of Pareto optimal solutions in $path^*$}
      \label{fig:4g}
    \end{center}
  \end{minipage}
  \begin{minipage}{0.325\textwidth}
    \begin{center}
      \includegraphics[width=1.0\textwidth,clip]{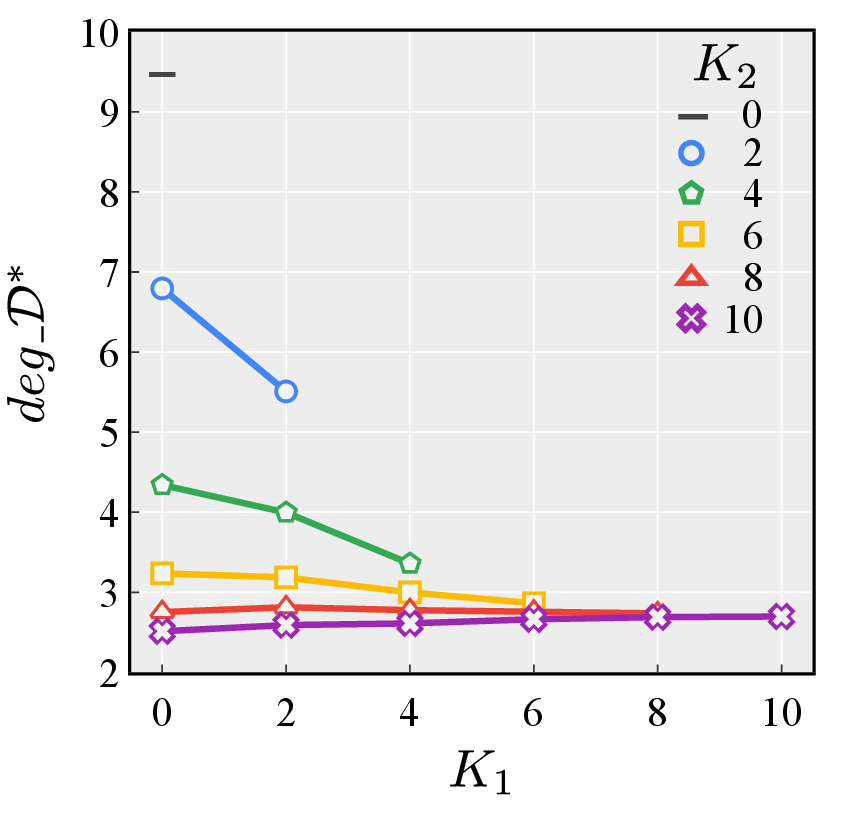}
      \\\vspace{-0.1cm}
      \subcaption{The degree of $D^*$.\\~}
      \label{fig:4h}
    \end{center}
  \end{minipage}
  \begin{minipage}{0.325\textwidth}
    \begin{center}
      \includegraphics[width=1.0\textwidth,clip]{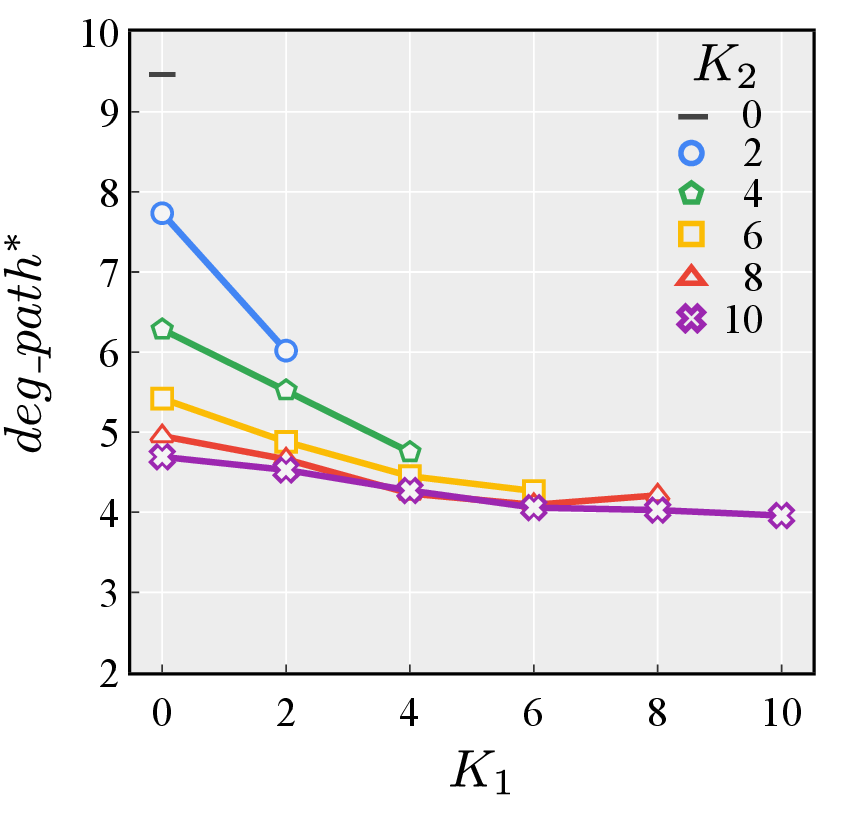}
      \\\vspace{-0.1cm}
      \subcaption{The degree of $path^*$.\\~}
      \label{fig:4i}
    \end{center}
  \end{minipage}
  \caption{The metrics of multi-objective landscapes by combining $NK$-landscapes with different difficulties. The color and shape of markers indicate the maximum $K$ of the $NK$-landscape in the $MNK$-landscape.}
  \label{fig:4}
\end{figure*}

The metrics values in \Tabref{tab:2} for the $1,770$ instances of the $MNK$-landscape are illustrated in \Figref{fig:4}. The average value for all instances corresponding to each parameter combination $\bm{K}$ is plotted. In each figure, the vertical axis is the metric value, the horizontal axis is the number of co-variables $K_1$ of the function $f_1$, and the six plots are respectively different numbers of co-variables $K_2$. Since the results for $\bm{K}=(K_1, K_2)$ and $\bm{K}=(K_2, K_1)$ are identical, the results are plotted only when $K_1 \leq K_2$.

\Figref{fig:4a} shows that the Pareto local optimal solution set $PL$ increases in the increase of $K_2$, which is the maximum number of co-variables between two $K$ values. \Figref{fig:4b} shows how many $PL$-domains the Pareto local optimal solutions are divided into. As with the number of Pareto local optimal solutions, $PL$-domains are affected by $K_2$ but do not increase monotonically. \Figref{fig:4c} shows the existing probability of the optimal $PL$-domain $D^*$ that contains global optima of both target and helper problems. Except for the combination $\bm{K}=(0, 2)$, we see the optimal $PL$-domain exists in more than $85^\%$ of instances. \Figref{fig:4d} and \Figref{fig:4e} respectively show the size of the optimal $PL$-domain and the percentage of the Pareto local optimal solutions within the optimal $PL$-domain among all Pareto local optimal solutions. The optimal $PL$-domain and the number of the Pareto local optimal solutions become larger as $K_2$ is higher. These results indicate that the higher $K_1$ and $K_2$ are, the more Pareto local optimal solutions are generated, resulting in more significant optimal $PL$-domains. Increasing the size of the optimal $PL$-domain makes reaching the domain easier from outside. Still, it also makes searching for global optimum within the domain more challenging.

Since the optimal $PL$-domain involves two global optima of the target and the helper problems, the shortest path $path^*$ between them connecting the Pareto local optimal solutions exists. \Figref{fig:4f} shows the length of $path^*$. The higher $K_2$ is, the longer $path^*$ is. That is, the number of transition steps between two global optima of the target and the helper problems via Pareto local optimal solutions increases with $K_2$ even in the same optimal $PL$-domain.

\Figref{fig:4g} shows the ratio of Pareto optimal solutions in $path^*$. The higher $K_2$ is, the smaller the ratio in the shortest path. That is, the Pareto local optimal solutions connecting the two global optima is not necessarily the Pareto optimal solution. 

\begin{figure*}[t]
  \begin{minipage}{0.475\textwidth}
    \begin{center}
      \includegraphics[width=1.0\textwidth,clip]{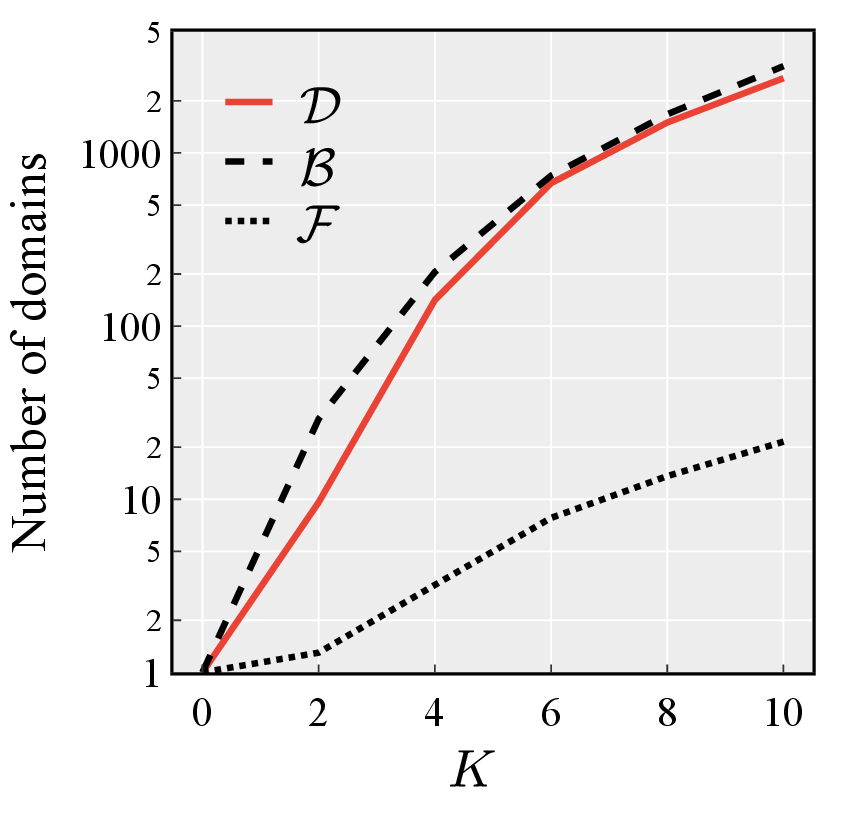}
      \\\vspace{-0.1cm}
      \subcaption{The Number of basins $\mathcal{B}$, funnels $\mathcal{F}$, and $PL$-domains $\mathcal{D}$.\\}
      \label{fig:5a}
    \end{center}
  \end{minipage}
  \begin{minipage}{0.475\textwidth}
    \begin{center}
      \includegraphics[width=1.0\textwidth,clip]{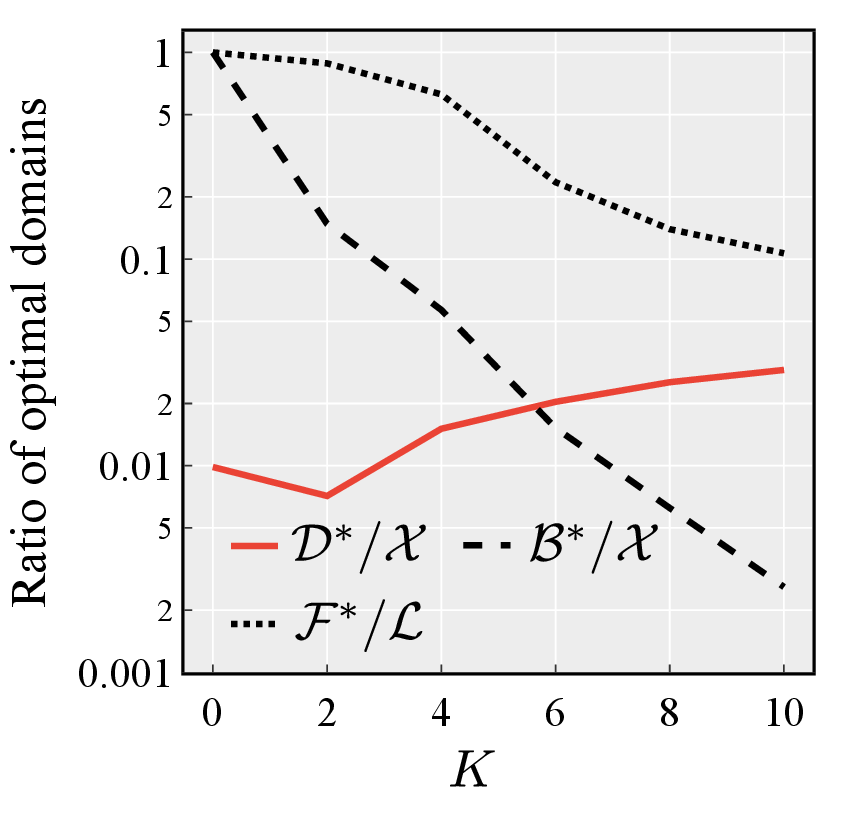}
      \\\vspace{-0.1cm}
      \subcaption{The relative size of the optimal basin $\mathcal{B}^*/\mathcal{X}$, funnels $\mathcal{F}^*/\mathcal{L}$, and $PL$-domains $\mathcal{D}^*/\mathcal{X}$ to search space.}
      \label{fig:5b}
    \end{center}
  \end{minipage}
  \caption{Comparison of single-objective landscapes and multi-objectivized landscapes with unimodal helper functions.}
  \label{fig:5}
\end{figure*}

\Figref{fig:4h} and \Figref{fig:4i} respectively show the average degrees in the optimal $PL$-domain $D^*$ and $path^*$. As $K_2$ increases, the degree in $D^*$ decreases, which means that the connection becomes sparse. The average degree in $path^*$ tends to be similar to that in $D^*$. However, we see the tendency that the average degree in $path^*$ is higher than the one $D^*$. This result suggests that $path^*$ is highly centered in the optimal $PL$-domain. Graph features would be useful to find the appropriate trajectory in $D^*$.

\Figref{fig:5} finally compares the number and size of basins, funnels for $60$ instances of $NK$-landscape, and optimal $PL$-domains for $600$ instances of $NK$-landscape with $K_1=0$. We see the number of basins and $PL$-domains increase exponentially with increasing $K$ or $K_2$. This result indicates that a single local optimum forms a single $PL$-domain. We see the funnel increase but more slowly than the number of $PL$-domain. Local optimum forms a funnel rather than a completely random distribution. As the number of basins and funnels increases, the size of the optimal ones becomes relatively small.
In contrast, the optimal $PL$-domain is slightly affected by $K$ and grows in size as $K$ increases. Small $D^*$ makes an exploration within $D^*$ easier, and the larger $D^*$ makes finding $D^*$ in search space easier. While conventional optimization algorithms search the optimal funnel or basin from the search space, the proposed MOLS explores within the optimal $PL$-domain from the known global optimum of the helper function. MOLS is considered most effective in small and dense connected $D^*$.

\subsection{Algorithm Benchmarks}
\begin{figure}[t]
    \begin{center}
    \includegraphics[width=0.93\linewidth]{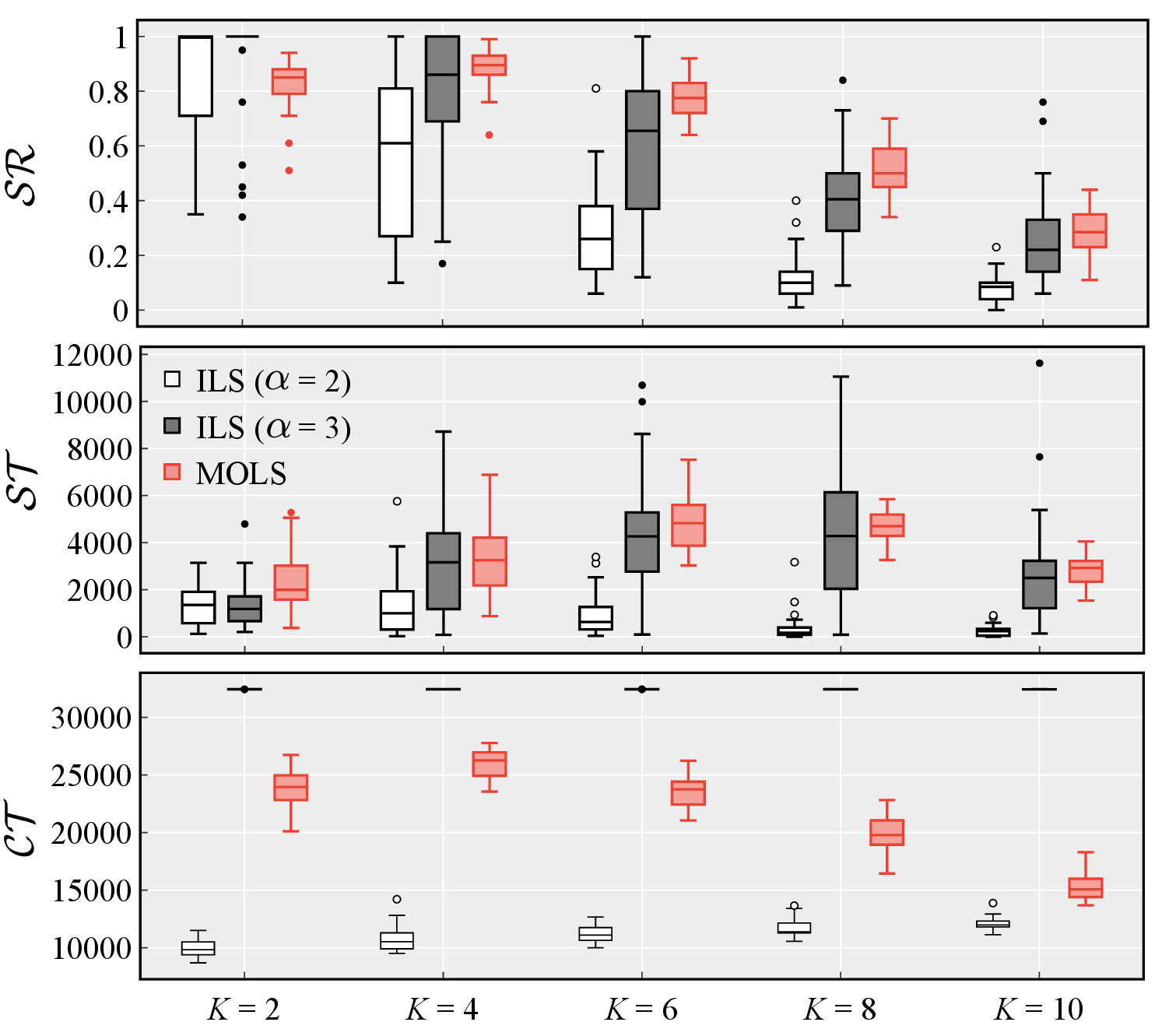}
    \caption{Search performances on target $NK$-landscape problems with $N=18, K\in \{2, 4, 6, 8, 10\}$.}
    \label{fig:6}
    \end{center}
\end{figure}

\Tabref{tab:4} and \Figref{fig:6} respectively show the average search performance values and the boxplots of search performance values with their distribution. From the results on $\mathcal{SR}$, the success rate of finding global optima, we see that the proposed MOLS performs better than the conventional ILSs for all $K$ except for $K=2$. This performance deterioration in $\bm{K}=(0, 2)$ instances is caused by the low probability of $D^*$ organized, as shown in \Figref{fig:4c}. This is because the proposed MOLS is an algorithm that searches within $D^*$, and if $D^*$ does not exist, the search fails to reach the global optimum of the target problem. From \Figref{fig:5}, we see that instances of the $NK$-landscape with $K=2$ have a large funnel of nearly $90^\%$ of the search space $X$. If the optimal funnel can be easily searched, as in the case of $NK$-landscape with $K=2$, there is a risk that multi-objectivization makes the target function more difficult. 
\begin{table*}[t!]
  \caption{Average search performance values}
  \label{tab:4}
  \begin{center}
  \resizebox{\textwidth}{!}
  {%
    \begin{tabular}{l|l|l|l|l|l|l}
    \hline 
    Metrics                    & Algorithms & $K=2$              & $K=4$              & $K=6$              & $K=8$              & $K=10$             \\ \hline \hline
    \multirow{3}{*}{$\mathcal{SR}$}     
    & Conventional ILS ($\alpha=2$)  & $0.835$            & $0.561$            & $0.289$            & $0.115$            & $0.079$            \\
    & Conventional ILS ($\alpha=3$)  & $\bm{0.915}$       & $0.772$            & $0.586$            & $0.410$            & $0.265$            \\
    & Proposed MOLS                 & $0.825$            & $\bm{0.883}$       & $\bm{0.771}$       & $\bm{0.519}$       & $\bm{0.282}$       \\ \hline
    \multirow{3}{*}{$\mathcal{ST}$}     
    & Conventional ILS ($\alpha=2$)  & $1.306 \times10^3$ & $1.408 \times10^3$ & $0.941 \times10^3$ & $0.381 \times10^3$ & $0.260 \times10^3$ \\
    & Conventional ILS ($\alpha=3$)  & $1.356 \times10^3$ & $3.247 \times10^3$ & $4.236 \times10^3$ & $4.315 \times10^3$ & $2.722 \times10^3$ \\
    & Proposed MOLS                 & $2.357 \times10^3$ & $3.322 \times10^3$ & $4.904 \times10^3$ & $4.654 \times10^3$ & $2.832 \times10^3$ \\ \hline
    \multirow{3}{*}{$\mathcal{CR}$} 
    & Conventional ILS ($\alpha=2$)  & $0.991 \times10^4$ & $1.074 \times10^4$ & $1.119 \times10^4$ & $1.171 \times10^4$ & $1.207 \times10^4$ \\
    & Conventional ILS ($\alpha=3$)  & $3.240 \times10^4$ & $3.240 \times10^4$ & $3.240 \times10^4$ & $3.240 \times10^4$ & $3.240 \times10^4$ \\
    & Proposed MOLS                 & $2.372 \times10^4$ & $2.592 \times10^4$ & $2.354 \times10^4$ & $1.992 \times10^4$ & $1.536 \times10^4$ \\ \hline
    \end{tabular}
  }
  \end{center}
\end{table*}
On the other hand, when $K$ is high, the success rate is only about $30^\%$, even though $D^*$ is more than $85^\%$ likely to exist. This result indicates that the proposed MOLS cannot explore the entire $D^*$. One reason the search performance of the proposed MOLS deteriorates as $K$ increases would be due to the connectivity of $D^*$. From \Figref{fig:4h}, we can expect that the higher $K$ of the target function is, the more sparse the $D^*$ becomes. Due to the nature of the proposed MOLS, which cannot transition back to a solution once searched, it is likely to converge on the terminal part of $D^*$.

From the results on $\mathcal{ST}$, the number of function evaluations when the global optimum is found, we see the proposed MOLS shows the highest values for all $K$. The range of $\mathcal{ST}$ for MOLS was within the range of $\mathcal{ST}$ for ILS with $\alpha=3$ except when $K=2$, and the proposed MOLS shows smaller variances of $\mathcal{ST}$ and more stable than ILSs. From the results on $\mathcal{CT}$, the number of function evaluations at the end of the search, ILS with $\alpha=3$ shows the highest values for all $K$. This is because the number of perturbations required to determine convergence increases as the ILS parameter $\alpha$ increases. In contrast, the termination condition of the proposed MOLS is whether or not a new Pareto local optimal solution exists in the neighborhood. Although it does not directly refer to the objective function value of the target function, the search can be terminated before the upper bound of function evaluation. The faster convergence with increasing $K$ may be due to the sparse connection of $D^*$. MOLS needs a mechanism that prevents it from falling into a dead end on a sparsely connected domain.

\section{Conclusions}
This paper investigated the effect of multi-objectivization on single-objective $NK$-landscape problems with multi-funnel structures. First, we used instances of $NK$-landscape problems with different $K$ and conducted the landscape analysis to quantitatively observe the landscape changes of the target problem by adding helper problems. Experimental results showed that the global optima of the target and the helper problems exist within one domain consisting of the Pareto local optimal solutions in many problem cases. The size of the bridging domain grew as the target function's global multimodality increased. This trend contrasted with existing concepts such as basin of attraction and funnels.
On the other hand, the connections within the bridging domain became sparse and inversely proportional to its size. In addition, we proposed a search algorithm named MOLS that searched for the global optimum of the target problem from the global optimum of an artificially generated helper problem via Pareto local optimal solutions. The proposed MOLS showed a better success rate of finding the target problem's global optimum than the conventional ILS in $NK$-landscape with multiple funnels. This result suggests that the multi-objectivization relaxed the problem difficulty of the target function due to the multiple funnel structure.

The results in this paper are limited to relatively small-scale $NK$-landscape problems. Also, the proposed MOLS is intuitive but consists of simple operators, and there is room for improvement in the update and termination phases. Experimental results suggested that graph features such as degree would be utilized in the search. In future work, we will improve the proposed search algorithm and analyze landscapes with a broad range of problems to examine the effectiveness of multi-objectivization.

\bibliographystyle{splncs04}
\bibliography{reference}

\end{document}